\documentclass[letterpaper, 10 pt, conference]{ieeeconf}

\IEEEoverridecommandlockouts                              

\usepackage{graphics} 
\usepackage{amsmath} 
\usepackage{amssymb}  
\usepackage{algorithm}%
\usepackage{algpseudocode}%
\usepackage{subcaption}
\usepackage[most]{tcolorbox}
\usepackage{tabularx}
\usepackage{multirow}
\usepackage{adjustbox}
\usepackage{float}
\tcbuselibrary{raster,skins,breakable,theorems}

\title{\LARGE \bf
Hierarchical LLM-Based Multi-Agent Framework \\
with Prompt Optimization for Multi-Robot Task Planning
}

\author{Tomoya Kawabe$^{1}$ and Rin Takano$^{1}$
\thanks{$^{1}$T. Kawabe and R. Takano are with Data Science
Laboratories, NEC Corporation, 1753, Shimonumabe, Nakahara-ku,
Kawasaki, Kanagawa, 211-8666, Japan.
{\tt\small \{tomoya-kawabe, rin\_takano\}@nec.com}}%
\thanks{\textcopyright~2026 IEEE. Personal use of this material
is permitted. Permission from IEEE must be obtained for all
other uses, in any current or future media, including
reprinting/republishing this material for advertising or
promotional purposes, creating new collective works, for
resale or redistribution to servers or lists, or reuse of
any copyrighted component of this work in other works.}%
}

\begin{document}


\maketitle

\thispagestyle{empty}
\pagestyle{empty}


\begin{abstract}

    Multi-robot task planning requires decomposing natural-language instructions into executable actions for heterogeneous robot teams. Conventional Planning Domain Definition Language (PDDL) planners provide rigorous guarantees but struggle to handle ambiguous or long-horizon missions, while large language models (LLMs) can interpret instructions and propose plans but may hallucinate or produce infeasible actions. We present a hierarchical multi-agent LLM-based planner with prompt optimization: an upper layer decomposes tasks and assigns them to lower-layer agents, which generate PDDL problems solved by a classical planner. When plans fail, the system applies TextGrad-inspired textual-gradient updates to optimize each agent's prompt and thereby improve planning accuracy. In addition, meta-prompts are learned and shared across agents within the same layer, enabling efficient prompt optimization in multi-agent settings. On the MAT-THOR benchmark, our planner achieves success rates of 0.95 on compound tasks, 0.84 on complex tasks, and 0.60 on vague tasks, improving over the previous state-of-the-art LaMMA-P by 2, 7, and 15 percentage points respectively. An ablation study shows that the hierarchical structure, prompt optimization, and meta-prompt sharing contribute roughly +59, +37, and +4 percentage points to the overall success rate.

\end{abstract}


\section{Introduction}

Multi-robot task planning has become an essential capability for robotics applications in household assistance, warehouse automation, and disaster response~\cite{background1, background2}. These scenarios typically involve heterogeneous robots with distinct capabilities, where high-level missions must be decomposed into executable subtasks and efficiently allocated among team members. Conventional planning approaches, such as symbolic planners based on the Planning Domain Definition Language (PDDL)~\cite{pddl}, provide formal correctness guarantees. However, they require precise problem specifications and scale poorly when applied to long-horizon, ambiguous, or dynamically changing tasks.

Advances in large language models (LLMs) have demonstrated remarkable potential to bridge this gap. LLMs are strong at interpreting natural language instructions, applying commonsense reasoning, and generating structured outputs such as action sequences or PDDL descriptions. This has enabled a new class of frameworks that leverage LLMs for robotic task planning. Unlike classical planners, LLM-based approaches can flexibly handle underspecified or ambiguous commands, decompose long-horizon tasks into manageable subtasks, and integrate external knowledge for reasoning. However, they also suffer from inherent limitations: hallucinations, logical inconsistencies, and a lack of formal guarantees often result in plans that are suboptimal or even infeasible.

\begin{figure}[t]
  \centering
  \hspace*{0cm}\includegraphics[width=0.47\textwidth]{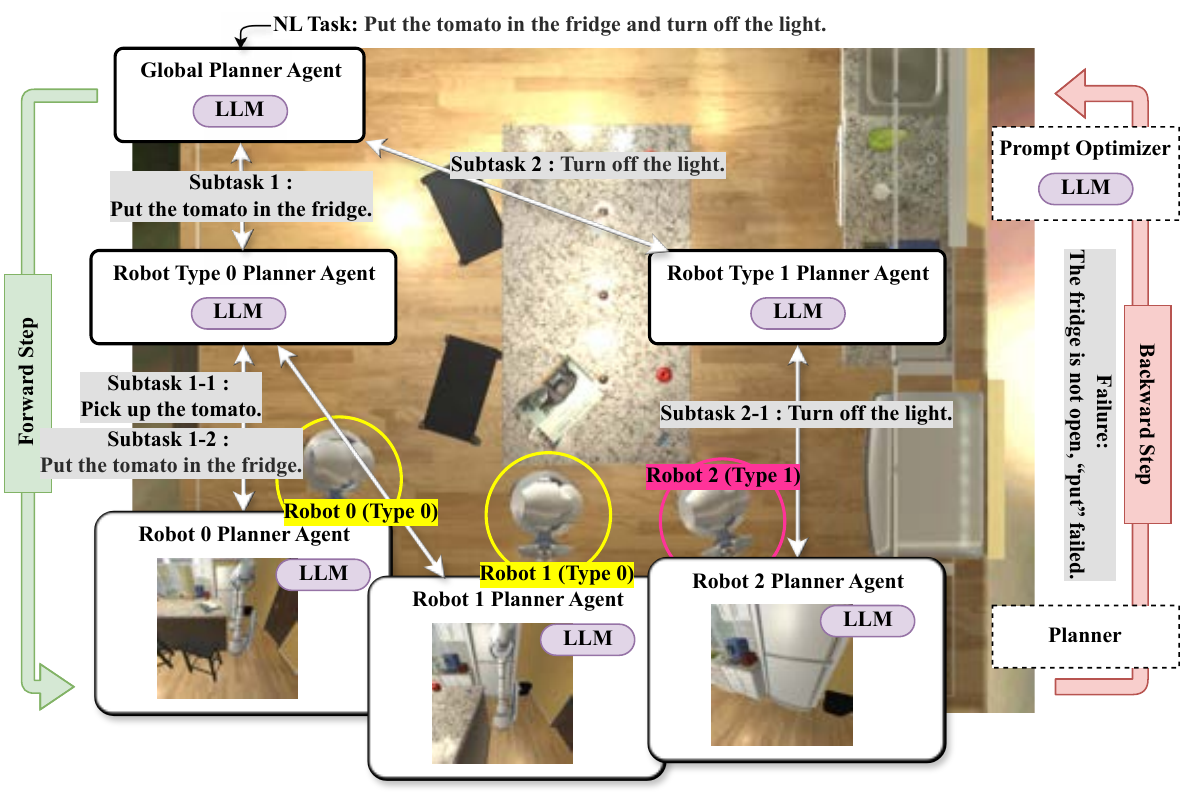}
  \caption{Multi-robot task planning in home environments. \textbf{Forward step}: Decomposing long-horizon tasks using a hierarchical LLM agent structure. \textbf{Feedback step}: Optimizing LLM agent prompts based on results verified by the planner.}
  \label{fig:intro}
\end{figure}

To address these limitations, hybrid frameworks that combine the generative reasoning of LLMs with the rigorous verification of classical or optimization-based planners have been proposed. 
While promising results, most existing methods have limitations such as relying on a single centralized LLM planner, which leads to computational bottlenecks and a lack of scalability as the number of robots or tasks increases.
Moreover, most existing methods operate as open-loop pipelines: once a plan is generated, execution failures are not propagated back to revise the decomposition or allocation policies.
Furthermore, the construction of feedback prompts strongly influences replanning quality, motivating automated prompt optimization rather than manual design.

To address these challenges, we propose a hierarchical multi-agent LLM-based planning system that distributes reasoning across multiple agents and that incorporates iterative prompt optimization to improve feasibility and scalability in complex multi-robot domains (see Figure~\ref{fig:intro}).

The key contributions of our work are as follows:
\begin{itemize}
    \item We propose a hierarchical multi-agent architecture that distributes task decomposition and allocation across layers of LLM agents, enabling scalability to large environments and long-horizon missions.
    \item We introduce a feedback-driven prompt optimization mechanism inspired by TextGrad~\cite{textgrad}, allowing agents to iteratively refine their prompts when execution failures occur.
    \item We further improve efficiency through meta-prompt sharing across homogeneous agents, leveraging ideas from meta-learning to accelerate adaptation in multi-agent settings.
\end{itemize}


\section{Related Work}

\subsection{Natural-language tasking with LLMs}
LLMs enable robots to be instructed in natural language (NL) while exploiting broad world knowledge for task interpretation and subtasking. SayPlan grounds NL instructions in large multi-room environments via 3D scene graphs to produce plans executable by navigation/manipulation stacks~\cite{sayplan}. SMART-LLM utilizes staged prompting with a single LLM to decompose and allocate tasks to heterogeneous robots~\cite{smartllm}. These systems demonstrate flexible NL tasking but also expose typical failure modes of monolithic LLM planning—hallucinated preconditions, inconsistent dependency structures, and degraded reasoning on long horizons—stemming from a single model carrying decomposition and allocation within a limited context.
More recently, Wang et al.~\cite{added1} integrate augmented scene graphs with LLMs to generate LTL-based task sequences for cross-regional multi-robot environments and select optimal plans via a heuristic function. While their approach strengthens spatial reasoning for task allocation, it operates as an open-loop pipeline without iterative prompt refinement from execution feedback.

\subsection{Hybrid LLM + classical/optimization planners}
To improve executability and correctness, hybrid approaches delegate search/verification to structured planners while using LLMs for NL understanding and problem shaping. LLM+P translates NL problems into PDDL, invokes a classical planner, and verbalizes the resulting plan back into NL~\cite{LLMP}. DELTA decomposes long-horizon goals into subgoals and solves each with a planner, improving the success and efficiency in complex scenes~\cite{delta}. Optimization- and constraint-centric variants similarly integrate LLM reasoning with formal solvers, e.g., linear programming in LiP-LLM~\cite{lipllm}, STL/TAMP translation in AutoTAMP~\cite{autotamp}, and constraint extraction/compilation in CaStL~\cite{castl}. While these hybrids curb hallucinations and raise executability, most pipelines centralize task decomposition and problem construction in a single LLM; as task horizons and environment scale increase—especially in multi-robot settings—this centralization strains context length and reduces planning fidelity.

\subsection{From centralized to distributed reasoning: multi-LLM agent systems}
To address scale, systems distribute reasoning across multiple LLM agents with explicit roles and interfaces. LaMMA-P instantiates role-specialized LLM modules (e.g., precondition identification, task allocation, problem generation, validation assistance) coupled with a PDDL planner, reporting state-of-the-art results on long-horizon, multi-robot household tasks~\cite{lammap}. RoCo and HMAS-II assign an LLM to each robot and coordinates subtask allocation via inter-agent dialogue before invoking a multi-arm motion planner~\cite{roco,HMAS2}. Cognitive-loop designs such as LLaMAR structure planning, acting, correction, and verification for multi-robot teams without relying on a single centralized LLM~\cite{llamar}. Division of labor alleviates single-model context limits and supports heterogeneous skills, but many pipelines remain largely unidirectional: once PDDL (or a validated plan) is produced, there are limited mechanisms to propagate failures upstream to revise decomposition/allocation policies or shared guidance.
Addressing reliability from a different angle, Wang et al.~\cite{added2} apply conformal prediction to distributed LLM-based action selection, achieving guaranteed mission success rates while mitigating hallucinations at decision time---in contrast to our post-hoc prompt optimization driven by planner-verified feedback.

The trajectory progresses from single-LLM NL tasking (flexible but fragile), to hybrid LLM+planner pipelines (feasible but centralized), to multi-LLM agent architectures (scalable via specialization). A remaining gap is a feedback-driven mechanism that connects planner-verified outcomes back to the prompts that generated them, including shared updates within layers of agents. Our work targets this gap by combining a hierarchical multi-agent design with iterative, planner-informed prompt updates and meta-prompt sharing.


\section{Multi-Robot Planning Problem Formulation}

Long-horizon tasks for heterogeneous robot teams require reasoning over the joint abilities of multiple robots, the environment state, and the order in which actions must be executed. Classic planners formalize a planning problem as a tuple that includes a set of actions and a transition function, but they usually assume a single robot and do not directly account for the allocation of tasks among multiple robots. Following work on language-model-driven planning, we cast our setting as a cooperative Multi-Robot Planning task. This section introduces the notation for the robotic planning problem; the LLM-based reasoning agents and hierarchy are defined later in Section~\ref{sec:agent-arch}.

\subsection{Multi-Robot planning problem definition}
\label{sec:map-def}

We define the \textbf{robot set} $R = \{r_1,\dots,r_N\}$, which contains the physical robots that execute actions in the environment. Robots belong to distinct \textbf{types}, which determine their skill sets. Let $\mathcal{T}$ be the finite set of types, and let $\mathsf{type}:R\to\mathcal{T}$ assign each robot to a type. Types correspond to PDDL domains: robots of the same type share the same PDDL domain and therefore the same skill primitives and action operators. We denote the domain associated with type $\tau$ by $D_{\tau}$, and we denote the set of skills available to robots of type $\tau$ by $\mathsf{cap}(\tau)\subseteq \Sigma$, where $\Sigma$ is the whole set of skill primitives.

For clarity, we define $N := |R|$ to be the number of robots and $M := |\mathcal{T}|$ to be the number of types of robots. We will use these quantities when describing algorithmic complexity and hyper-parameters later in the paper.

Formally, our multi-robot task planning problem is specified by the tuple
\begin{equation}
\Pi = \langle R,\,\mathcal{T},\,\mathsf{type},\,\mathsf{cap},\,P,\,I,\,G,\,{\{A^{r}\}}_{r\in R}\rangle,
\end{equation}
where:
\begin{itemize}
    \item \textbf{Robots $R$.} The set of physical robots available to perform the task.
    \item \textbf{Types $\mathcal{T}$.} Each type $\tau\in\mathcal{T}$ defines a PDDL domain with predicates and operators. The mapping $\mathsf{type}$ assigns every robot $r$ to a type $\mathsf{type}(r)$.
    \item \textbf{Skills $\mathsf{cap}$.} The function $\mathsf{cap}:\mathcal{T}\to 2^\Sigma$ maps each type to the set of skills (capabilities) it can perform. Robots of the same type have the same set of skills. For instance, a mobile base may have skills $\{\mathrm{move}\}$, while a manipulator may have skills $\{\mathrm{move},\mathrm{pickup},\mathrm{putdown}\}$. Throughout this paper we use the term ``skills'' for these sets and write $\mathsf{cap}(\tau)$ for the skills of type $\tau$.
    \item \textbf{State atoms $P$.} A finite set of propositional atoms (fluents) describing the world state, such as object locations, robot locations and grasping status.
    \item \textbf{Initial state $I$.} A subset $I\subseteq P$ describing the state before planning begins.
    \item \textbf{Goal condition $G$.} A subset $G\subseteq P$ describing the desired state after task completion.
    \item \textbf{Action sets $A^{r}$.} For each robot $r$, the action set $A^{r}$ consists of parameterized actions built from the skills in $\mathsf{cap}(\mathsf{type}(r))$. Each action $a\in A^{r}$ is defined by a pair $\langle \mathrm{pre}(a),\mathrm{eff}(a)\rangle$, where $\mathrm{pre}(a)\subseteq P$ is the set of preconditions and where $\mathrm{eff}(a)\subseteq P\cup\{\lnot p\mid p\in P\}$ is the set of effects. The PDDL representation of actions uses human-readable names; for example, the \texttt{PickupObject} operator has preconditions \texttt{(at-location ?object ?location)} and \texttt{(at ?robot ?location)} and effects \texttt{(holding ?robot ?object)}.
\end{itemize}

The \textbf{transition function} $\delta(s,a)$ applies an action $a$ to a state $s\subseteq P$ by adding its positive effects and by removing its negative effects:
\begin{equation}
\delta(s,a) = \bigl( s\cup\{ p \mid p\in\mathrm{eff}(a) \} \bigr) \setminus \{ p \mid \lnot p\in\mathrm{eff}(a) \}.
\end{equation}

An action $a\in A^{r}$ is applicable in state $s$ if and only if $\mathrm{pre}(a)\subseteq s$. A \textbf{multi-robot plan}~\cite{MAP_survey} for $\Pi$ is an ordered set of action instances 
\begin{equation}
\Pi_g = \langle \Delta,\prec\rangle,
\end{equation} 
where $\Delta \subseteq \bigcup_{r\in R} A^{r}$ is a set of ground actions (each assigned to a specific robot) and where $\prec$ is a strict partial order of encoding precedence constraints. If $a\prec b$, then action $a$ must precede $b$. Actions that are not related by $\prec$ can be executed concurrently by different robots. Final state that can be obtained by executing $\Pi_g$ from $I$ must satisfy the goal condition $G$.

The \textbf{cost} of a sequential plan is defined as the number of action steps, denoted as $|\Pi|$. When actions execute concurrently, the plan cost (or makespan) is the number of parallel time steps required. Our objective is to find a plan with minimal cost subject to feasibility
.


\section{Proposed Method: Hierarchical LLM-Based Multi-Agent Planning}

In this section we present a hierarchical MAP framework that combines the high-level reasoning and language understanding of large language models (LLMs) with the rigor of classical planning. A team of LLM-driven agents is organized into a hierarchy. Upper layers decompose a natural-language task into subtasks and assign them to downstream agents, while leaf agents translate their assigned subtasks into formal PDDL problems and use a classical planner to generate executable plans. After each iteration, agents refine their prompts through textual-gradient feedback and share meta-prompt updates with their peers to improve efficiency and robustness.

\subsection{Hierarchical LLM Agent Architecture and Notation}
\label{sec:agent-arch}

We introduce the \textbf{agent set} $\mathcal{E}$, which contains the logical LLM-based reasoning agents distinct from the physical robots $R$. To capture the hierarchical structure used in our framework, we partition $\mathcal{E}$ into layers indexed by $l\in \{0,1,\dots,L-1\}$. Let $\mathcal{E}_l = \{ E_{l,0}, E_{l,1}, \dots, E_{l,|\mathcal{E}_l|-1}\}$ denote the set of agents in layer $l$, and define the overall agent set as the union of layer-specific sets:
\begin{equation}
\mathcal{E} = \bigcup_{l=0}^{L-1} \mathcal{E}_l.
\end{equation}

In particular, agents in layer $0$ form the highest level of abstraction (global reasoning), while agents in deeper layers refine tasks and eventually generate PDDL plans. Each agent $E_{l,i}$ maintains:
\begin{itemize}
  \item a \textbf{task} $\Psi(E_{l,i})$, represented as natural-language or structured text;
  \item a \textbf{prompt} $\theta_{E_{l,i}}$, which conditions its LLM behavior;
  \item a layer-shared \textbf{meta-prompt} $\hat{\theta}_l$.
\end{itemize}
When $l<L-1$, $E_{l,i}$ performs task decomposition for the next layer; when $l=L-1$, $E_{l,i}$ performs PDDL generation. This hierarchical design distributes reasoning across agents, avoiding the scalability bottleneck of a single monolithic LLM planner and enabling parallelism.

\begin{figure*}[t]
  \centering
  \hspace*{-1.75cm}\includegraphics[width=1.1\textwidth]{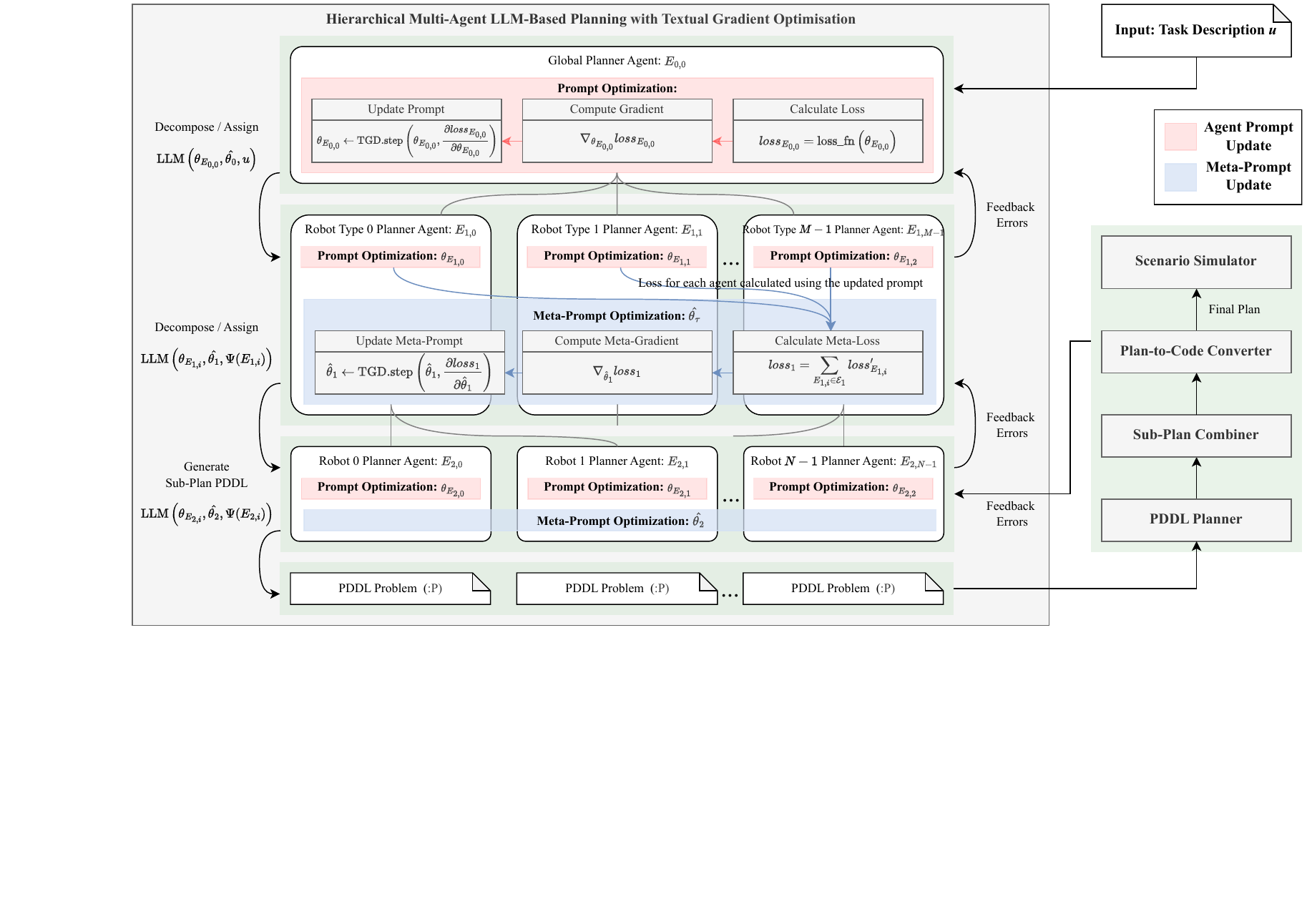}
  \vskip-4.5cm
  \caption{Overview of the hierarchical MAP framework.
  The red flows indicate \textbf{prompt optimization for each agent},  
  while the blue flows denote \textbf{meta-prompt optimization across layers}.}
  \label{fig:overview}
\end{figure*}

\begin{table}[t]
\centering
\caption{Algorithm overview}
\begin{tabularx}{\columnwidth}{p{2.3cm}X}
\hline
\textbf{Step} & \textbf{Description} \\ \hline
\emph{Inputs} &
Task instruction $u$; agents $\mathcal{E}$ with prompts $\theta_{E_{l,i}}$ and meta-prompts $\hat{\theta}_l$; maximum iterations $K_{\max}$. \\
Data structures &
Plan list $\Phi$, task map $\Psi$, set of sub-plans (PDDL specifications). \\
Initialization &
Set iteration counter $k\!\leftarrow\!0$, $\Phi\!\leftarrow\![E_{0,0}]$, $\Psi(E_{0,0})\!\leftarrow\!u$. \\
Top-down reasoning &
For each layer $l=0$ to $L-1$:  
each agent $E_{l,i}$ decomposes its task into subtasks;  
leaf agents generate PDDL problems. \\
Classical planning &
Each PDDL problem is solved with a planner; success/failure recorded. \\
Replanning &
Failed agents climb hierarchy by LLM decision (``self'' vs. ``parent''). \\
Prompt optimization &
Agents update prompts via textual gradients;  
layers update meta-prompts. \\
Termination &
If all sub-plans succeed, output them;  
else increment $k$ and repeat until success or $K_{\max}$. \\ \hline
\end{tabularx}
\end{table}

\algrenewcommand\algorithmicindent{0.8em}

\begin{algorithm}[tbp]
\caption{Hierarchical Multi-Agent LLM-Based Planning with Textual-Gradient Optimization}
\label{algo:hmalop}
\begin{algorithmic}[1]\small
\Require Task $u$; layered agent set $\mathcal{E}=\bigcup_{l=0}^{L-1}\mathcal{E}_l$; agent prompts $\{\theta_{E_{l,i}}\}$; meta-prompts $\{\hat{\theta}_l\}$; max iterations $K_{\max}$
\State $k \gets 0$ \Comment{iteration counter}
\State $\Phi \gets [E_{0,0}]$ \Comment{root agent in plan list}
\State $\Psi(E_{0,0}) \gets u$ \Comment{assign root task}

\While{true}
  \If{$k = K_{\max}$}
     \State \Return failure \Comment{stop if iteration limit reached}
  \EndIf
  \State $\mathcal{S} \gets \emptyset$ \Comment{candidate sub-plans (PDDL specs)}

  \For{$l \gets 0$ \textbf{to} $L-1$} \Comment{top-down reasoning}
    \State $S_l \gets [\,E \in \Phi \mid E \in \mathcal{E}_l\,]$
    \ForAll{$E_{l,i} \in S_l$}
      \If{$l < L-1$} \Comment{intermediate layer}
        \State $\{u_e\} \gets LLM(\theta_{E_{l,i}},\hat{\theta}_l,\Psi(E_{l,i}))$
        \ForAll{$E_{l+1,e} \in \mathcal{E}_{l+1}$}
          \State $\Phi.\mathrm{append}(E_{l+1,e})$
          \State $\Psi(E_{l+1,e}) \gets u_e$
        \EndFor
      \Else \Comment{leaf layer: PDDL generation}
        \State $\mathrm{pddl\_spec} \gets LLM(\theta_{E_{l,i}},\hat{\theta}_l,\Psi(E_{l,i}))$
        \State $\mathcal{S} \gets \mathcal{S} \cup \{\mathrm{pddl\_spec}\}$
      \EndIf
    \EndFor
  \EndFor

  \ForAll{$sp \in \mathcal{S}$} \Comment{classical planning \& validation}
    \State $(\,\mathrm{plan},\,E_{\mathrm{src}}) \gets \mathrm{PDDL\_Planning\_And\_Validate}(sp)$
    \If{planning/validation fails for $sp$}
      \State $E' \gets E_{\mathrm{src}}$
      \While{true} \Comment{decide where to replan: self vs parent}
        \State $\mathrm{decision} \gets LLM(\breve{\theta}_{E'},sp)$
        \If{$\mathrm{decision}=\text{``self''} \lor E'=E_{0,0}$}
          \State $\Phi.\mathrm{append}(E')$; \textbf{break}
        \Else
          \State $E' \gets \mathrm{ParentOf}(E')$
        \EndIf
      \EndWhile
      \State $(\Phi, \{\theta_{E_{l,i}}\}, \{\hat{\theta}_l\}) \gets \Call{promptUpdate}{\Phi, \{\theta_{E_{l,i}}\}, \{\hat{\theta}_l\}}$ \Comment{Algorithm\,\ref{algo:promptupdate}}
      \State $k \gets k+1$; \textbf{continue} \Comment{start next outer iteration}
    \EndIf
  \EndFor

  \State \Return $\mathcal{S}$ \Comment{all sub-plans validated}
\EndWhile
\end{algorithmic}
\end{algorithm}

\begin{algorithm}[tbp]
\caption{PromptUpdate: Agent- and Layer-level Prompt Updates (MAML~\cite{maml} inspired)}
\label{algo:promptupdate}
\begin{algorithmic}[1]\small
\Function{PromptUpdate}{$\Phi, \{\theta_{E_{l,i}}\}, \{\hat{\theta}_l\}$}
  \State $\mathrm{RemoveChildrenFrom}\,\Phi()$ \Comment{prune obsolete child agents}

  \Comment{\emph{(A) Agent-level inner updates: }\ $\theta^{(k)} \to \theta^{(k+1)}$}
  \ForAll{$E_{l,i} \in \Phi$}
    \State $\mathrm{loss}^{\mathrm{pre}}_{E_{l,i}} \gets \mathrm{loss\_fn}(\theta_{E_{l,i}})$ \Comment{compute loss}
    \State $g_{E_{l,i}} \gets \nabla_{\theta_{E_{l,i}}}\,\mathrm{loss}^{\mathrm{pre}}_{E_{l,i}}$ \Comment{gradient from feedback}
    \State $\theta_{E_{l,i}} \gets \mathrm{TGD.step}(\theta_{E_{l,i}}, g_{E_{l,i}})$ \Comment{update agent prompt}
    \State $\mathrm{loss}^{\mathrm{post}}_{E_{l,i}} \gets \mathrm{loss\_fn}(\theta_{E_{l,i}})$ \Comment{re-evaluate \emph{after} update}
  \EndFor

  \Comment{\emph{(B) Layer-level outer updates}}
  \For{$l \gets 1$ \textbf{to} $L-1$}
    \State $\text{loss}_l \gets \mathcal{A}_l\!\left(\{\mathrm{loss}^{\mathrm{post}}_{E_{l,i}} \mid E_{l,i}\in\mathcal{E}_l\}\right)$
    \Comment{meta-loss}
    \State $\hat{g}_l \gets \nabla_{\hat{\theta}_l}\,\text{loss}_l$ \Comment{meta-gradient}
    \State $\hat{\theta}_l \gets \mathrm{TGD.step}(\hat{\theta}_l, \hat{g}_l)$ \Comment{update meta-prompt}
  \EndFor
  \State \Return $(\Phi, \{\theta_{E_{l,i}}\}, \{\hat{\theta}_l\})$
\EndFunction
\end{algorithmic}
\end{algorithm}

\subsection{Overview of Algorithm}

Assume a user issues a high-level instruction $u$ to accomplish a long-horizon task. The framework maintains a set of LLM agents $\mathcal{E}=\bigcup_{l=0}^{L-1}\mathcal{E}_l$, partitioned by layer~$l$ (Section\,\ref{sec:map-def}). Each agent $E_{l,i}$ has its own prompt $\theta_{E_{l,i}}$ and shares a \emph{meta-prompt} $\hat{\theta}_l$ with peers in its layer. Iterations continue until all leaf PDDL plans are validated or until a maximum number of iterations $K_{\max}$ is reached.

Algorithm~\ref{algo:hmalop} depicts the procedure at a high level.  The outer loop (lines~4-41) controls the iterations.  At each iteration it clears the set of \emph{candidate sub-plans} and performs a top-down pass over the hierarchy (lines~9-23):

\begin{itemize}
  \item \textbf{Task decomposition} (lines~12-17): For every agent $E_{l,i}$ not at the leaf layer, the LLM uses its prompt $\theta_{E_{l,i}}$, the layer meta-prompt $\hat{\theta}_l$ and its assigned task $\Psi(E_{l,i})$ to generate a set of subtasks for the agents in layer~$l+1$.  The plan list $\Phi$ is updated to include these child agents, and their tasks are stored in $\Psi$.
  \item \textbf{PDDL generation} (lines~19-20): For every leaf agent ($l=L-1$), the LLM produces a domain and problem in PDDL form.  These \emph{sub-plans} are collected for validation.
\end{itemize}

After the top-down pass, each candidate sub-plan is passed to a classical planner/validator.  We use an off-the-shelf PDDL planner such as Fast\,Downward~\cite{Helmert2006}. If all the plans succeed, they are returned as the solution.  Otherwise, we identify the agent whose plan failed and climb the hierarchy until either the agent chooses to revise its plan or the root is reached.  Similar to the two-step optimization pipeline proposed by Shen et\,al.~\cite{DBLP} for multi-agent systems, our system then performs targeted prompt updates: each failing agent (and its ancestors) calculates a textual loss using its LLM, back-propagates a textual gradient to its prompt, and updates its prompt via a TextGrad-style optimizer~\cite{textgrad}.  Finally, each layer aggregates losses across agents and applies a meta-prompt update (lines~8-11 in Algorithm~\ref{algo:promptupdate}), analogous to meta-learning across homogeneous agents.

Algorithm~\ref{algo:hmalop} describes the high-level procedure, while
Figure~\ref{fig:overview} provides a visual overview of the hierarchical agent
interactions and optimization flows.

\subsection{Hierarchical Multi-Agent Architecture}

The hierarchy $\mathcal{E}$ organizes reasoning agents into layers $l=0,\dots,L-1$. Agents in higher layers decompose tasks, while leaf agents generate PDDL problems. Each agent $E_{l,i}$ maintains (i) a task $\Psi(E_{l,i})$, (ii) a prompt $\theta_{E_{l,i}}$, and (iii) an optional layer-shared meta-prompt $\hat{\theta}_l$. 

In our experiments we adopted a three-layer hierarchy: a \textbf{global planner layer} interprets the user instruction, a \textbf{type layer} distributes subtasks to robot types based on their skills, and a \textbf{robot layer} generates PDDL for individual robots. This decomposition avoids the scalability bottleneck of a single monolithic LLM planner and enables parallel execution across heterogeneous agents.

\subsection{Classical Planning and Validation}

Once the leaf agents generate PDDL specifications, the proposed method invokes a classical planner to compute executable plans.  We utilize the Fast\,Downward planner with the search and heuristic settings recommended in prior work~\cite{Helmert2006}.  For each PDDL problem $sp$, a plan in returned or failure is indicated.  If a plan exists, it is validated to ensure that applying the actions from the initial state achieves the goal.  This validation step is critical: language models can produce syntactically valid but logically inconsistent PDDL that fails at runtime~\cite{Helmert2006}.  Only after all sub-plans succeed at planning and validation does the method terminate successfully.

If some sub-plans fail, the system must decide where to replan.  Each failing agent $E$ consults a \textbf{replanning prompt} $\breve{\theta}_{E}$ to determine whether it should attempt to revise its sub-task (decision~``self'') or request a higher-level agent to rethink the decomposition (decision~``parent'').  This strategy is inspired by hierarchical error propagation in multi-agent systems~\cite{DBLP}.

\subsection{Textual-Gradient Prompt Optimization}

The quality of LLM outputs depends heavily on the prompts. To improve reliability over multiple iterations, our framework incorporates a \emph{textual-gradient optimization} mechanism~\cite{textgrad}. After each iteration, every agent $E_{l,i}$ receives feedback indicating how its output failed. These errors are summarized as a textual loss $\text{loss}_{E_{l,i}}$. Each agent's prompt $\theta_{E_{l,i}}$ and each layer's meta-prompt $\hat{\theta}_l$ are then updated using textual-gradient descent, as outlined in Algorithm~\ref{algo:hmalop} and executed in Algorithm~\ref{algo:promptupdate}.

\paragraph{Agent-level update.}
For each agent $E_{l,i}$, natural-language feedback from the classical planner and downstream agents is mapped to a textual loss via the same loss function used throughout Algorithm~\ref{algo:promptupdate}. With the iteration index managed in Algorithm~\ref{algo:hmalop}, the agent-level inner update at iteration $k$ is:
\begin{align}
  \mathrm{loss}^{(k)}_{E_{l,i}} \;&:=\; \mathrm{loss\_fn}\!\bigl(\theta^{(k)}_{E_{l,i}}\bigr), \label{eq:agent-pre-loss}\\
  g^{(k)}_{E_{l,i}} \;&:=\; \nabla_{\theta_{E_{l,i}}}\,\mathrm{loss}^{(k)}_{E_{l,i}}, \label{eq:agent-grad}\\
  \theta^{(k+1)}_{E_{l,i}} \;&:=\; \mathrm{TGD.step}\!\bigl(\theta^{(k)}_{E_{l,i}},\, g^{(k)}_{E_{l,i}}\bigr). \label{eq:agent-tgd}
\end{align}
Here, $\mathrm{TGD.step}$ denotes a textual gradient descent step that \emph{rewrites} the prompt $\theta_{E_{l,i}}$ by applying a small set of ranked edit operations suggested by the LLM (e.g., adding clarifying constraints or reordering checks). No auxiliary parameters are introduced beyond the prompt itself and its textual ``gradient'' $g^{(k)}_{E_{l,i}}$. Equations~\eqref{eq:agent-pre-loss}--\eqref{eq:agent-tgd} correspond to the agent-level inner loop in Algorithm~\ref{algo:promptupdate}.

\paragraph{Layer-level update.}
After the agent-level inner updates (Algorithm~\ref{algo:promptupdate}, lines 4--10), we evaluate each updated agent prompt with the same loss function:
\begin{equation}
\text{loss}_{E_{l,i}}^{(k+1)} \;:=\; \mathrm{loss\_fn}\!\bigl(\theta_{E_{l,i}}^{(k+1)}\bigr),
\qquad E_{l,i}\in\mathcal{E}_l.
\label{eq:post-agent-loss}
\end{equation}
To obtain a single layer-wise signal, we aggregate the per-agent textual losses using an LLM-based operator $\mathcal{A}_l$ that \emph{deduplicates overlapping elements, normalizes phrasing, and consolidates common edits} across agents in layer $l$:
\begin{equation}
\text{loss}_l^{(k+1)} \;=\;
\mathcal{A}_l\!\Big(\big\{\text{loss}_{E_{l,i}}^{(k+1)}\big\}_{E_{l,i}\in\mathcal{E}_l}\Big).
\label{eq:layer-agg}
\end{equation}
We then compute a textual ``gradient'' of this meta-loss with respect to the layer meta-prompt and apply one textual-gradient descent step:
\begin{align}
\hat{g}_l^{(k+1)} \;&=\; \nabla_{\hat{\theta}_l}\,\text{loss}_l^{(k+1)},\\
\hat{\theta}_l^{(k+1)} \;&=\; \mathrm{TGD.step}\!\bigl(\hat{\theta}_l^{(k)},\,\hat{g}_l^{(k+1)}\bigr).
\label{eq:meta-step}
\end{align}

\noindent
This realizes a MAML~\cite{maml} inspired bilevel structure in discrete text space: inner (agent) adaptation produces \(\theta_{E_{l,i}}^{(k+1)}\) and post-update losses \eqref{eq:post-agent-loss}, while the outer (layer) update aggregates them via \eqref{eq:layer-agg} and updates the shared meta-prompt via \eqref{eq:meta-step}. Notationally, the iteration index \(k\) is maintained in Algorithm~\ref{algo:hmalop}; \textsc{PromptUpdate} (Algorithm~\ref{algo:promptupdate}) takes the current \(\Phi^{(k)}, \{\theta^{(k)}\}, \{\hat{\theta}^{(k)}\}\) and returns \(\Phi^{(k+1)}, \{\theta^{(k+1)}\}, \{\hat{\theta}^{(k+1)}\}\) without requiring \(k\) as an explicit argument. The operator \(\mathcal{A}_l\) is implemented as a prompt that consolidates per-agent textual losses into a single layer-level objective and a ranked set of candidate edits; only the consolidated objective is used to compute \(\hat{g}_l^{(k+1)}\).

\subsection{Example Multi-Agent Configuration}

A concrete example of the hierarchical architecture clarifies its operation.  Consider a household assistance domain.  The top layer contains a \textbf{global planner agent} that receives a natural-language instruction such as ``tidy the living room and prepare tea.''  This agent decomposes the instruction into subtasks and assigns them to agents in the second layer, where each \textbf{type agent} corresponds to a category of robots (e.g., mobile base, manipulator).  Type agents further refine their assigned subtasks and assign them to individual \textbf{robot agents} in the third layer.  For instance, the mobile-base agent might generate a PDDL problem requiring movement to specific locations, while the manipulator agent produces a PDDL problem involving ``pickup'' and ``putdown'' actions.

To illustrate PDDL generation, suppose the manipulator agent is assigned the task ``place the laptop on the desk.''  It constructs a PDDL domain and problem following standard syntax.  The domain includes an operator such as
\begin{tcolorbox}[
  breakable,
  colback=white, colframe=black,
  listing only, 
  listing options={basicstyle=\ttfamily\small, breaklines=true}]
(:action PickupObject
 :parameters (?r - robot ?o - object ?l - location)
 :precondition (and (at ?o ?l) (at ?r ?l))
 :effect (and (holding ?r ?o) (not (at ?o ?l))))
\end{tcolorbox}
and a corresponding operator for \texttt{PutdownObject}.  The problem file defines the initial state (object and robot locations) and the goal \texttt{(holding ?r ?o)} or \texttt{(at ?o desk)} as appropriate.  These PDDL specifications are passed to the classical planner, which returns a sequence of actions.  By varying the decomposition and agent assignment, the same framework can support other domains such as warehouse automation or disaster response.


\section{Numerical Experiments}
\label{sec:numerical-experiments}

We empirically evaluate our hierarchical multi-agent planner on the MAT-THOR benchmark for long-horizon household tasks, originally proposed in LaMMA-P~\cite{lammap}. MAT-THOR extends the AI2-THOR~\cite{ai2thor} simulator and provides 70 tasks across five floor plans with increasing complexity and ambiguous instructions. Each task is annotated with a natural-language instruction, a ground-truth PDDL domain, and a goal condition. To account for simulator stochasticity, we execute each task with five random initializations and report the averages.

\subsection{Task categories and dataset}
\label{sec:dataset}

MAT-THOR organizes its 70 tasks into three categories based on their structure. \emph{Compound} tasks contain two to four largely independent subtasks and can be executed in parallel.  \emph{Complex} tasks consist of six or more subtasks with causal dependencies, often requiring robots with complementary skills.  \emph{Vague command} tasks deliberately omit crucial details, forcing the agent to infer missing information from context. The benchmark includes 30 compound tasks, 20 complex tasks, and 20 vague commands, providing detailed specifications of initial states, robot skill sets, and success conditions.  These tasks support the evaluation of task decomposition, allocation, and execution efficiency by heterogeneous teams of two to four robots.

\begin{table*}[t]
\centering
\caption{Performance comparison on MAT-THOR.  Higher values are better for all
metrics.}
\label{tab:overall}
\begin{tabular}{l|cccc|cccc|cccc}
\hline
\multirow{2}{*}{Method}
  & \multicolumn{4}{c|}{Compound}
  & \multicolumn{4}{c|}{Complex}
  & \multicolumn{4}{c}{Vague}\\
\cline{2-13}
  & SR & GCR & RU & Eff & SR & GCR & RU & Eff & SR & GCR & RU & Eff\\
\hline
CoT~\cite{CoT} (GPT-4o) & 
0.32 & 0.40 & 0.72 & 0.59 & 
0.00 & 0.12 & 0.47 & 0.38 & 
0.00 & 0.00 & 0.00 & 0.00 \\
SMART-LLM~\cite{smartllm} (GPT-4o) & 
0.70 & 0.82 & 0.78 & 0.64 & 
0.20 & 0.33 & 0.65 & 0.56 & 
0.06 & 0.42 & 0.68 & 0.42 \\
LaMMA-P~\cite{lammap} (GPT-4o) &
0.93 & 0.94 & 0.91 & \textbf{0.90} & 
0.77 & 0.83 & 0.87 & 0.67 & 
0.45 & 0.48 & 0.71 & 0.65 \\
\textbf{Ours (GPT-4o)} & 
\textbf{0.95} & \textbf{0.95} & \textbf{1.00} & \textbf{0.90} & 
\textbf{0.84} & \textbf{0.84} & \textbf{1.00} & \textbf{0.75} & 
\textbf{0.60} & \textbf{0.60} & \textbf{1.00} & \textbf{0.75} \\
\hline
\end{tabular}
\end{table*}

\subsection{Evaluation metrics and baselines}
\label{sec:metrics}

Following LaMMA-P~\cite{lammap}, we assess plan quality using four metrics: success rate \textbf{(SR)}, goal condition recall \textbf{(GCR)}, robot utilization \textbf{(RU)}, and efficiency \textbf{(Eff)}. 
The SR is the fraction of tasks in which all goal conditions are achieved. 
The GCR is the ratio of achieved goal atoms to the ground-truth goal set. 
The RU measures how efficiently robot actions are used by comparing the total transition count to the ground truth.  
The Eff captures the temporal efficiency of the plan as the ratio between the makespan of the generated plan and the ground-truth solution; both the RU and Eff are computed only on successful executions.

We compare our method against three baselines. Chain-of-Thought (CoT)~\cite{CoT} prompting uses GPT-4o to directly translate the instruction into action sequences for each robot.  
SMART-LLM~\cite{smartllm} decomposes and allocates subtasks sequentially and relies on a motion planner for trajectory generation.  
LaMMA-P~\cite{lammap} combines LLM reasoning with PDDL planning and currently achieves state-of-the-art performance on MAT-THOR.  
For a fair comparison, all methods utilize the same PDDL planner (Fast\,Downward with the LAMA heuristic)~\cite{Helmert2006} and are evaluated in identical simulation environments.  
The CoT~\cite{CoT} and SMART-LLM~\cite{smartllm} baselines are run using GPT-4o, following the configuration described in the LaMMA-P~\cite{lammap}.

\subsection{Implementation details and hyperparameters}

All the LLM agents in our hierarchy --global, type, and robot-- are implemented using GPT-4o. Each agent starts from an initial prompt that conveys its role: the global agent is instructed to summarize the user's instruction and decompose it into high-level subtasks; type-level agents allocate these subtasks to robots based on capability descriptions; robot-level agents generate PDDL domains and problems describing their assigned sub-task. A shared meta-prompt per layer provides additional context about typical household tasks and the available skill types. The maximum number of prompt-optimization iterations is set to $K_{\max}=5$. The execution is simulated in AI2-THOR~\cite{ai2thor}. All of the methods share identical simulation settings to ensure comparability.

\subsection{Case study: prompt optimization in practice}
\label{sec:case-study}


To illustrate how textual feedback improves performance, we consider a compound task from MAT-THOR: ``\emph{put the tomato in the fridge and turn off the room light.}'' In our initial run, the Robot Type~0 agent assigns the fridge task to Robot~0, which generates a PDDL problem. The classical planner succeeds at \texttt{Pickup} but fails at \texttt{Put} with a precondition error because the fridge has not been opened. The failing agent logs this feedback and uses the textual gradient mechanism to suggest inserting an \texttt{Open} action before any \texttt{Put} into a receptacle with open/close affordance. After updating its prompt, the agent adds a subtask assigning Robot~1 to open the fridge. The second iteration, however, fails with a path-planning error because Robot~1 remains in front of the fridge. A further prompt update instructs the robot to move to a non-blocking waypoint after opening, and the third iteration completes successfully. These small prompt edits, generalised through meta-prompt sharing, eliminate entire classes of failures without manual intervention. The supplementary video walks through this example in detail. Detailed examples of these prompt updates are provided in Appendix~\ref{app:prompt-updates}.

\subsection{Overall performance}
\label{sec:overall-performance}

Table\,\ref{tab:overall} compares our hierarchical planner with the
baselines on MAT-THOR. The numbers for CoT~\cite{CoT} and SMART-LLM~\cite{smartllm} are taken from
LaMMA-P's evaluation~\cite{lammap}. Our method achieves the
highest success rates across all task categories.  
For compound tasks, the success rate reaches~0.95, slightly higher than LaMMA-P's 0.93.  
On complex tasks, our method attains a success rate of 0.84, representing a clear improvement of 7 percentage points over LaMMA-P's 0.77 and much higher than SMART-LLM's 0.20.  
Even on vague command tasks, which are particularly challenging due to underspecified instructions, our approach achieves 0.60, substantially higher than LaMMA-P's 0.45 and far above other baselines.  
These results highlight that our hierarchical decomposition and feedback-driven prompt optimization yield significant gains in success rate, especially in the most demanding task categories.

\subsection{Ablation study}

To evaluate the contributions of each component, we perform an ablation study in which hierarchical decomposition (\textbf{H}), local prompt optimization (\textbf{P}), and meta-prompt sharing (\textbf{M}) are removed individually or in combination. The results are summarized in Table \ref{tab:ablation}.

Removing the hierarchy (\text{--H}) and relying on a single LLM agent causes the most severe degradation, dropping the success rate to 0.25 (-59.3\%). This highlights the critical role of hierarchical decomposition in limiting input length and enabling parallelism. Disabling both prompt and meta-prompt optimization (\text{--(P,\,M)}) reduces the success rate to 0.47 (-37.0\%), showing that feedback-driven prompt adaptation is essential for executability and recovery from planning errors. Removing only meta-prompt optimization (\text{--M}) has a smaller effect, lowering the success rate to 0.80 (-4.2\%), but still demonstrates the benefit of sharing prompt refinements across homogeneous agents to accelerate convergence.

We also compare the average computation time across variants. The full method requires 173 seconds on average. Removing the hierarchy (\text{--H}) shortens runtime to 140 seconds. Removing meta-prompt optimization (\text{--M}) reduces runtime further to 116 seconds. Disabling both prompt and meta-prompt optimization (\text{--(P,\,M)}) makes planning fastest at just 32 seconds, but executability and success rate suffer significantly. Overall, the hierarchy delivers the largest performance gain, prompt optimization enables significant recovery from mistakes, and meta-prompt sharing provides incremental improvements, while computation-time results highlight the trade-off between efficiency and reliability.

\begin{table}[t]
\centering
\caption{Ablation study on our method (GPT-4o). Values are averages over the entire dataset. Both SR (success rate) and Time are reported relative to the full method (H+P+M).}
\label{tab:ablation}
\begin{tabular}{lll}
\hline
Variant & SR & Time \\
\hline
Full method (H+P+M)                      & \textbf{0.84} & 173 s \\
\text{--H} (single-LLM agent)            & 0.25 (-59.3\%) & 140 s (-18.9\%) \\
\text{--M} (no meta-prompt optimization) & 0.80 (-4.2\%)  & 116 s (-33.1\%) \\
\text{--(P, M)} (no prompt optimization) & 0.47 (-37.0\%) & \textbf{ 32 s} (-81.3\%)  \\
\hline
\end{tabular}
\end{table}


\section{Conclusion}

We presented a hierarchical multi-agent LLM-based planning framework that integrates natural-language reasoning with classical PDDL planning and feedback-driven prompt optimization. The system improves scalability and reliability in multi-robot task planning by distributing tasks across layered agents and refining prompts through textual gradients. On the MAT-THOR benchmark, it outperforms prior LLM-based planners, with gains of up to 9.1\% on complex tasks and 33.3\% on vague commands. Ablation studies confirm that hierarchy and prompt optimization are critical to performance.

Limitations remain: the fixed hierarchy reduces adaptability, full observability is assumed, and prompt optimization still faces issues of convergence speed and stability. Future work will explore adaptive hierarchies, integration with perception in partially observable settings, and more robust optimization for real-world deployment.


\appendix

\section{Prompt update examples}
\label{app:prompt-updates}

The prompt update examples illustrate how textual gradients modify assignment prompts during the case study described in Section\,\ref{sec:case-study}.
At each iteration, a failure triggers
a suggestion to add a specific check or precondition.  These changes
accumulate, and because the meta-prompt is shared within each layer,
subsequent tasks benefit from the updated instructions.

\newtcolorbox{promptupdate}[2][]{%
  colback=white, colframe=black,
  fonttitle=\bfseries,
  title=Iteration #2,
  breakable,
  #1
}

\begin{promptupdate}{0}
\textbf{Before:}
\small{
``Decompose into subtasks as needed and assign them to robots. Hint:~\texttt{""}''
}

\textbf{After:}
\small{
    Append ``before putting the tomato into the fridge, it is necessary to open the fridge.''
    Meta-prompt updated to ``for any subtask that places into a receptacle with open/close affordance, insert \texttt{Open} before any \texttt{Put}.''
}
\end{promptupdate}

\begin{promptupdate}{1}
\textbf{Before:}
\small{Includes previous update}

\textbf{After:}
\small{Append ``after opening the fridge, move to a non-blocking waypoint to clear the doorway.''
Meta-prompt updated to ``append an egress action to a non-blocking waypoint to clear the doorway.''}
\end{promptupdate}

In subsequent iterations, no further updates were necessary; the
accumulated prompt modifications successfully eliminated the observed
failure modes. Agents of the same type avoided similar mistakes in future
tasks by propagating the modifications through the shared
meta-prompt.




\bibliographystyle{IEEEtran}
\bibliography{ref}

\end{document}